\documentclass[letterpaper, 10 pt, conference]{ieeeconf}  

\IEEEoverridecommandlockouts                              

\title{\LARGE \bf
Look Before You Leap: Socially Acceptable High-Speed Ground Robot Navigation in Crowded Hallways
}

\author{Lakshay Sharma$^{1}$ and Jonathan P. How$^{1}$
\thanks{$^{1}$Aerospace Controls Laboratory, Massachusetts Institute of Technology, 77 Massachusetts Ave., Cambridge, MA 02139, USA
        {\tt\small \{lakshays, jhow\}@mit.edu}}%
}

\usepackage{xcolor}
\usepackage{graphicx}
\usepackage{siunitx}
\usepackage{amsmath}
\usepackage{algorithm}
\usepackage{algpseudocode}
\usepackage{float}
\DeclareMathOperator*{\argmin}{arg\,min}

\begin{document}

\maketitle
\thispagestyle{empty}
\pagestyle{empty}

\begin{abstract}

To operate safely and efficiently, autonomous warehouse/delivery robots must be able to accomplish tasks while navigating in dynamic environments and handling the large uncertainties associated with the motions/behaviors of other robots and/or humans. A key scenario in such environments is the hallway problem, where robots must operate in the same narrow corridor as human traffic going in one or both directions. Traditionally, robot planners have tended to focus on socially acceptable behavior in the hallway scenario at the expense of performance. This paper proposes a planner that aims to address the consequent "robot freezing problem" in hallways by allowing for "peek-and-pass" maneuvers. We then go on to demonstrate in simulation how this planner improves robot time to goal without violating social norms. Finally, we show initial hardware demonstrations of this planner in the real world.

\end{abstract}

\section{Introduction}

Human-robot co-existence and collaboration is essential to expanding current automation capabilities. Social robot navigation, which comprises as planning robot behavior in a dynamic, uncertain environment co-occupied by humans is therefore of increasing importance. Increasing robot and human productivity in a shared workspace requires improvements in two key areas:

\begin{itemize}
    \item Average robot speed: improvements are needed to maximize robot productivity and minimize inter-stage latency
    \item Human comfort with robot behavior: with humans feeling uneasy or unsafe around robots in a professional environment, productivity of both agents drops
\end{itemize}

This leads to the realization that there is a need for high speed, safe and socially mobile robot motion planners that can safely negotiate space while obeying appropriate social politeness norms.

Now, it is usually noted that the two objectives above demonstrate tradeoffs against each other. Additionally, human comfort is a subjective metric which is hard to quantify accurately, leading to robots usually being isolated to prevent human injury in a workspace. Consequently, it is all too common for state of the art social navigation algorithms to pick one extreme or another: either planners are too aggressive, leading to rude and/or unpredictable behaviors which make nearby humans uneasy (less common), or planners are too conservative, leading to robots being stuck behind slow pedestrians and unable to complete their tasks efficiently (more common).

This paper therefore attempts to find a middle ground, by proposing a socially acceptable planner that solves the "robot freezing problem" of being stuck behind a slow human, by executing an overtaking maneuver. However, to do this safely, the robot must be careful to look before it leaps: peeking before passing to ensure safety for itself and any unseen human who might be occupying the unobserved space needed to pass.
\section{Related Work}

Social navigation is a rich field of study, and there have been many attempts at solving these issues. We can broadly categorize planners into two kinds: learning-based and optimization-based.

Learning-based planners such as \cite{chen2017decentralized} are rapidly emerging as a robust way to plan in social environments. However, we choose to focus on optimization-based planners such as \cite{truong2017toward}, \cite{singamaneni2021human} for our approach to solve the social navigation issues under consideration.

There has also been much work in the field of risk metrics. In particular, \cite{majumdar2020should} and \cite{feng2024safe} discuss risk metrics that should be used in robotics. In this paper, 

Recent work done in \cite{francis2023principles} mentions several key metrics using which human comfort can be measured, which are rapidly being adopted by the community. We use several of these metrics to quantify out planners against the chosen baselines.
\section{Pipeline Overview}

\begin{figure}[h]
\centering
\includegraphics[width=0.45\textwidth]{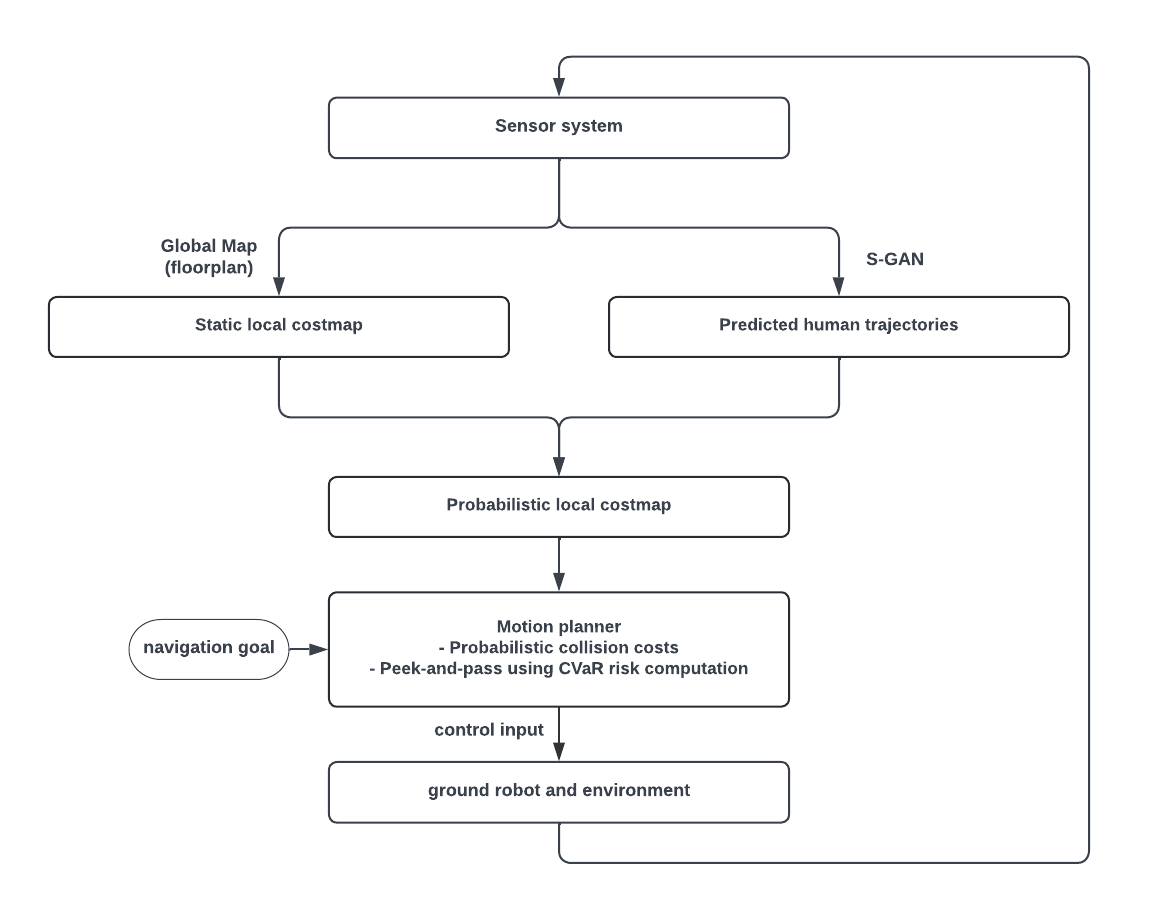}
\caption{Overview of the proposed planning pipeline. \label{fig:pipeline}}
\end{figure}

Our pipeline, outlined in Fig. \ref{fig:pipeline} consists of three levels: mapping (conversion of sensor data to a 2D costmap), prediction (predicting future human trajectories for dynamic obstacle avoidance), and planning (generation and execution of a motion plan given the costmap).

\subsection{Mapping}

The robot is assumed to have a global map of its environment so that walls, shelves and other permanent static obstacles are predetermined. Additionally, we take as given the following sensors: a 2D laser scanner, and two stereo depth cameras -- one pointing forwards and one pointing backwards. These sensors are then used to generate an occupancy map, with obstacle inflation then producing a 2D costmap for the instantaneous environment state.

\subsection{Prediction}

The 2D costmap generated above is not sufficient to plan in an environment with dynamic human agents. Therefore, we generate future human trajectories using an S-GAN based online predictor, similar to \cite{poddar2023crowd}. This creates a time-dependent distribution for occupancy probabilities, and therefore costs, that is overlaid on the costmap to be used by the planner.

\subsection{Planning}

Given the 2D costmap with static obstacles and distributions of positions, costs and occupancy probabilities for dynamic obstacles, this paper's key contribution is a novel risk-aware motion primitives-based robot motion planner that targets the robot freezing problem in hallways. By allowing for peek-and-pass maneuvers, we demonstrate improved times to goal without sacrificing social acceptability. The details of this planner are described below in Section IV.
\section{Planner}

With the costmap described in Sections III A and B above, we can proceed with generating a robust motion plan for the robot to execute. We consider here the hallway passing problem, described in Fig. \ref{fig:passing_scenario}. 

In a social robot navigation scenario, especially in a spatially narrow setting such as a hallway, human comfort with robot motion is critical. One of the most common ways that robots can induce social discomfort is by executing jerky motions and sharp turns that greatly reduce robot predictability. Therefore, we opt to use a motion primitives based approach similar to \cite{dharmadhikari2020motion}, since by generating plans that are guaranteed to respect vehicle dynamics they offer greater social comfort at high speeds than MPC-based methods. 

Aside from social acceptability, the key risk that we must account for when going fast in a hallway is that of humans approaching from the opposite direction in the unknown space created by a human right in front of the robot. To achieve this, we use the CVaR metric \cite{majumdar2020should} to predict the risk of future robot plans into unknown space.

\subsection{Problem Formulation}

We work with a differential drive ground robot operating in a 2D plane, subject to unicycle dynamics:

\begin{align}
    \begin{bmatrix}
        \dot{x}\\
        \dot{y}\\
        \dot{\theta}
    \end{bmatrix} = 
    \begin{bmatrix}
        \cos{\theta} && 0\\
        \sin{\theta} && 0\\
        0 && 1
    \end{bmatrix} \cdot
    \begin{bmatrix}
        v \\ \omega
    \end{bmatrix}
\end{align} \label{eq:unicycle}

where $x$ and $y$ are the coordinates of the robot in the $x-y$ plane it is operating in, and $\theta$ is the orientation of the robot with respect to a designated front. 

\begin{figure}[h]
    \centering
    \includegraphics[width=0.25\textwidth]{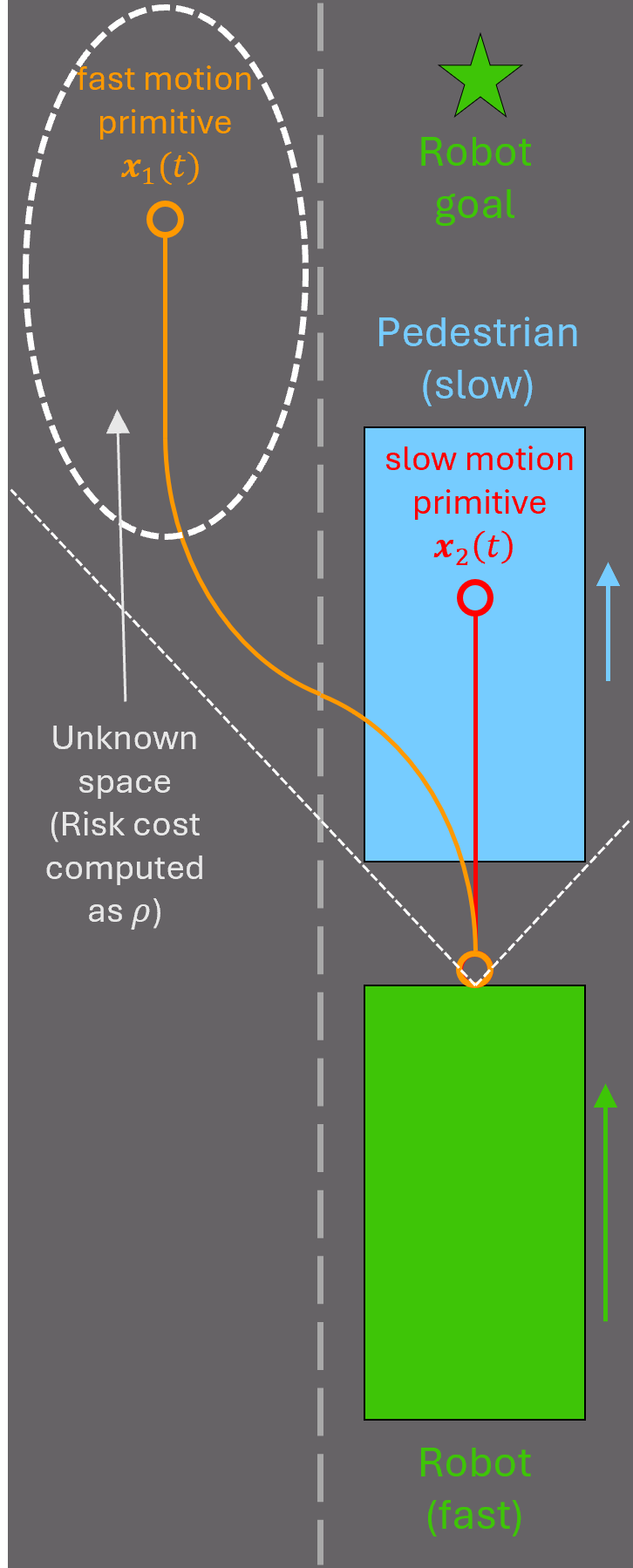}
    \caption{Hallway passing scenario for a fast ground robot stuck behind a slow human. \label{fig:passing_scenario}}
\end{figure}

Now consider the hallway passing scenario shown in Fig. \ref{fig:passing_scenario}. We first formulate an optimization problem, which when solved gives us a plan with the following properties:

\begin{itemize}
    \item High average speed (low time to goal) for the robot
    \item Obstacle avoidance, specifically:
    \begin{itemize}
        \item deterministic obstacle avoidance for static obstacles, and
        \item stochastic risk-based avoidance for dynamic obstacles (humans) 
    \end{itemize}
    \item Cautious "peek-and-go" behavior when the robot is blocked by a slow human
    \item Social acceptability of robot motion, particularly minimizing:
    \begin{itemize}
        \item personal space violations, and
        \item human trajectory interruptions (number of times a human has to change their path due to the robot cutting them off)
    \end{itemize}
\end{itemize}

To get these properties, the optimization problem to be solved is:
\begin{align}
    \label{eq:optimization_problem}
    \mathbf{x}^*&(t) = \argmin_{\mathbf{x}} J(\mathbf{x}(t),\mathbf{u})
\end{align}
where $J(\mathbf{x}(t),\mathbf{u})$ is a cost function that embodies the desired properties of goal-seeking, safe and socially acceptable exploration behavior outlined above.

\subsection{Cost Function}

Referencing Fig. \ref{fig:passing_scenario}, for a candidate trajectory $\mathbf{x}(t)$ obtained as a sequence of motion primitives, we then formulate a cost function for the planner to optimize as follows:

\begin{align}
    \label{eq:cost}
    J(\mathbf{x}&(t),\mathbf{u}) = c_1 \|\mathbf{x}(t_f) - \mathbf{x_g}\|^2 + c_2 p_{\text{coll}}(\mathbf{x}(t)) + c_3 \rho(\mathbf{x}(t)),\\
    \text{s.t. } & \mathbf{\dot{x}} = A\mathbf{x} + B\mathbf{u},\\
    & p_{\text{coll}}(\mathbf{x}(t)) = \sum _ {i = 0} ^k p_{\text{coll}}^{i}(\mathbf{x}(t)),\\
    & \rho(\mathbf{x}(t)) = \mathrm{max}(0, \mathrm{CVaR}_{\alpha_{\text{peek}}}(\boldsymbol{\Psi}(x_k(t), Y(t)) - \epsilon_{\text{peek}}),
\end{align}
where $c_1, c_2 \text{ and } c_3$ are tunable scalar cost parameters, $\alpha_{\text{peek}}$ and $\epsilon_{\text{peek}}$ are tunable risk parameters for the peeking maneuver, $t_f$ is the planning horizon of the planner, $\mathbf{x_g}$ is the goal state, $k$ is the number of humans/dynamic agents being tracked and predicted, $p_{\text{coll}}(\mathbf{x}(t))$ and $\rho(\mathbf{x}(t))$ are the risks of collision for a candidate trajectory $\mathbf{x}(t)$ for observable and unobservable obstacles, respectively. Note that $p_{\text{coll}}^0(\mathbf{x}(t))$ is the chance of collision with a static observable obstacle, calculable directly from the 2D occupancy map, whereas $p_{\text{coll}}^i(\mathbf{x}(t)) \text{ } \forall \text{ } i \neq 0$ is the chance of collision with the $i^{\text{th}}$ observed human/dynamic agent, calculated as a CVaR cost from the predicted trajectory distribution.

Focusing more on $\rho(\mathbf{x}(t))$, we first formulate a distribution $\boldsymbol{\Psi}$ representing the possible trajectories a human moving from an unobserved region of the map into the currently observable local map (denoted here as $Y(t)$). This distribution, sampled uniformly, is responsible for the emergence of peeking behavior rather than full-speed overtaking in the resultant plans. This is because a full-speed passing maneuver incurs two kinds of risk:

\begin{itemize}
    \item risk of collision with an unseen human on the passing side, and
    \item risk of collision with an unseen human in front of the human being passed.
\end{itemize}

Since the distribution $\boldsymbol{\Psi}$ captures both of these risks and $\rho(\mathbf{x}(t))$ penalizes them accordingly, our planner incentivizes peeking before passing, rather than risky overtaking at high speeds without sufficient information.
\section{Results}

The motion planning pipeline described thus far is tested against several other planners in simulation, and is also demonstrated on a real robot in a hallway.

\subsection{Preliminary Laned Simulation Results}

\begin{figure}[h]
    \centering
    \includegraphics[width=0.45\textwidth]{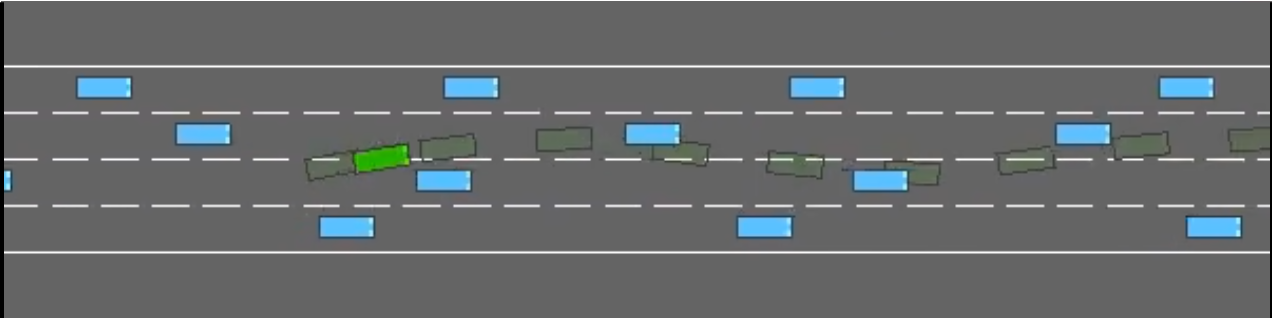}
    \caption{A laned simulation environment where a robot (green) navigates safely around several humans (blue) which are all moving left to right at varying speeds. \label{fig:laned_sim}}
\end{figure}

To first demonstrate that our planner improves performance over a conservative, non-peeking approach, we use a laned simulation environment called HighwayEnv \cite{highway-env} shown in Fig. \ref{fig:laned_sim}. This allows us to test the planner's performance in a well-structured environment. 

\begin{figure}[h]
    \centering
    \includegraphics[width=0.45\textwidth]{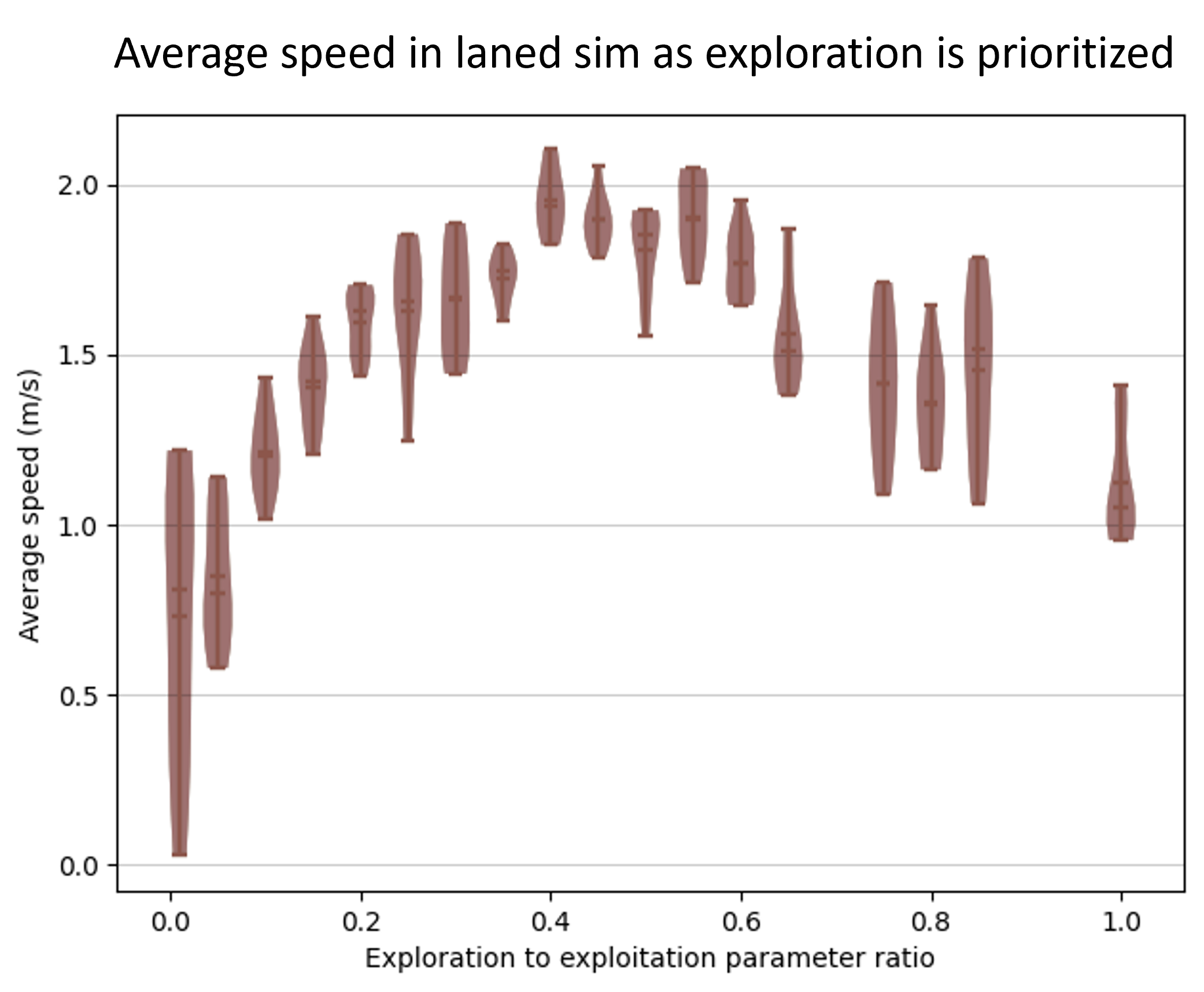}
    \caption{The laned simulation environment demonstrates that in a structured scenario, the proposed idea of "peek-and-go" successfully increases performance by actively sensing and clearing unknown space. \label{fig:laned_sim_results}}
\end{figure}

Using this environment to test a robot with a maximum speed of \qty{3}{m/s}, we see in Fig. \ref{fig:laned_sim_results} that as the exploration to exploitation parameter ratio for the planner (defined as $\frac{1}{c_1 c_3}$ using the notation in Eq. \ref{eq:cost}) increases, higher average speeds are achieved, until an optimal value, beyond which the robot loses a strong sense of attraction for the goal. In particular: 

\begin{itemize}
    \item Low exploration rates lead to low average speeds for the robot, with wide distributions of behavior that reflect reliance on lucky gaps
    \item High exploration rates show diminishing returns and even performance reduction, as the robot starts to emphasize uncertainty reduction over progress to the goal
    \item Optimal behavior is shown as a significant increase in speed at intermediate values of $\frac{1}{c_1 c_3}$ (exploration vs. exploitation)
\end{itemize}

These results show that the proposed idea of exploration/"peeking to pass" successfully increases robot performance without sacrificing safety. We also use this environment to tune the cost parameters. 

\subsection{Unlaned Photorealistic Simulation Results}

\begin{figure}[!h]
    \centering
    \begin{tabular}{cc}
        \includegraphics[height=0.175\textheight]{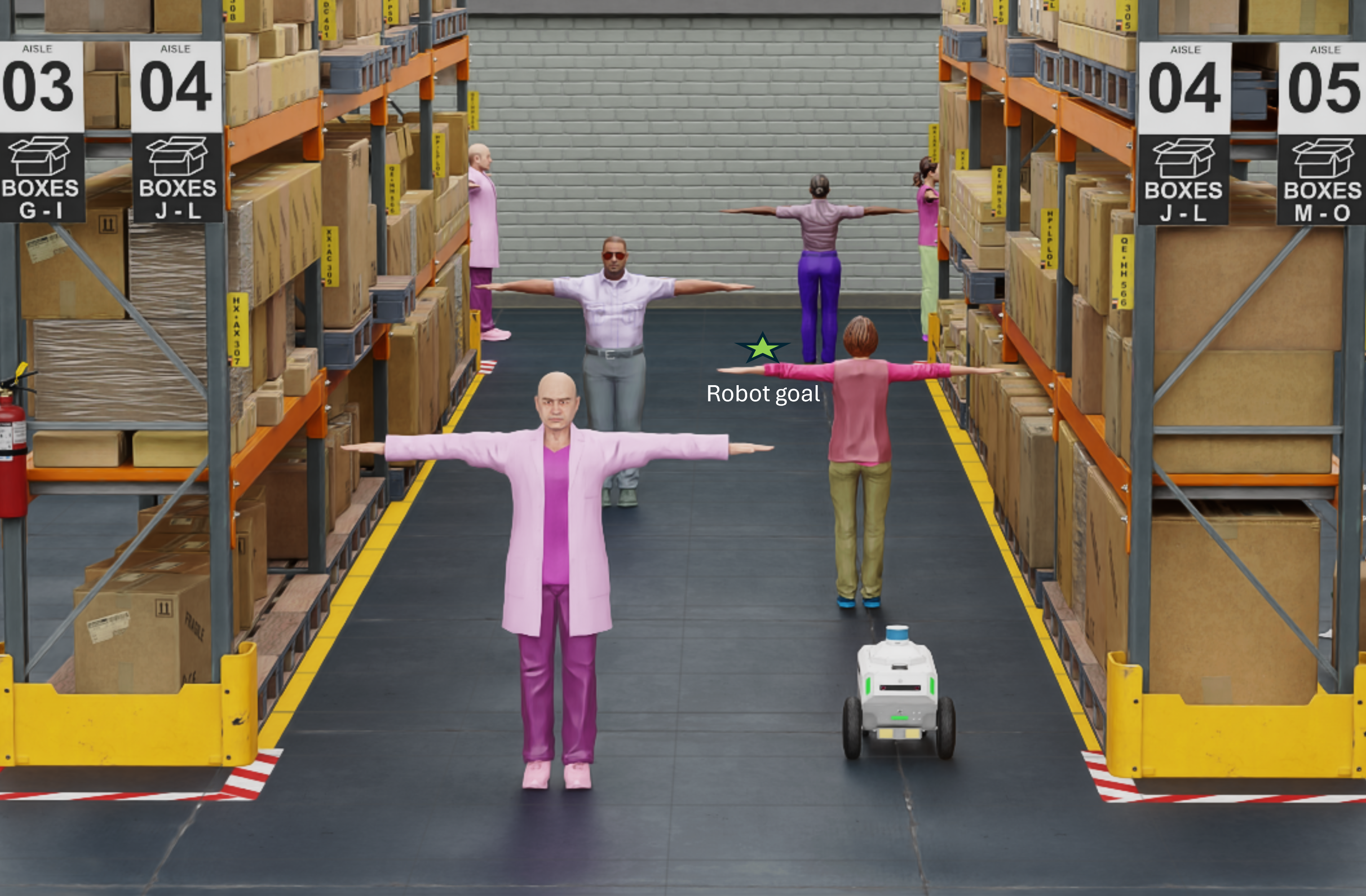} &  
        \includegraphics[height=0.175\textheight]{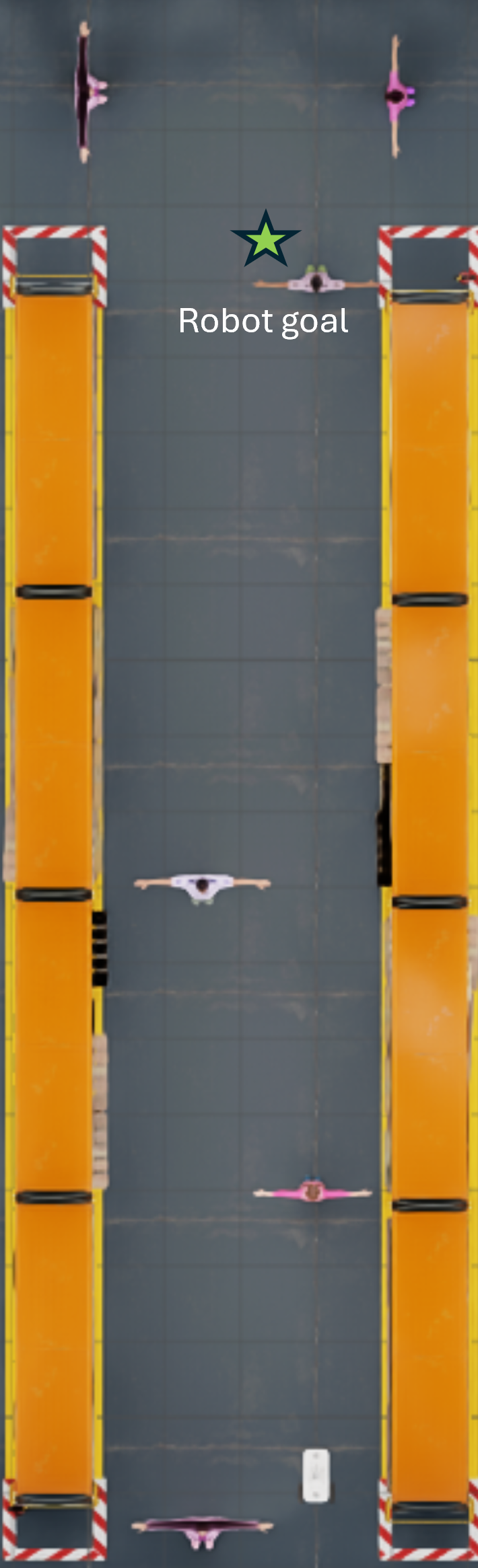} \\
        (a) & (b)
    \end{tabular}
    \caption{Unlaned \label{fig:sim_screenshot}}
\end{figure}

Having verified the performance gains of our approach above, we move to a photorealistic unlaned simulation in order to show detailed comparisons against baselines. We used NVIDIA Isaac Sim to recreate a crowded hallway environment with people walking both ways, as shown in Fig. \ref{fig:sim_screenshot}. The length of this hallway was varied between \qty{25}{m}, \qty{50}{m} and \qty{100}{m}, and the allowable human speeds (fixed during a single trial) range from \qty{0}{m/s} to \qty{3.5}{m/s} in increments of \qty{0.25}{m/s}. A differential drive robot with a top speed of \qty{3}{m/s} is spawned into this environment near one end of the hallway for each trial, and given a goal at the other end of the hallway.

\begin{figure*}[th]
    \centering
    \includegraphics[width=\textwidth]{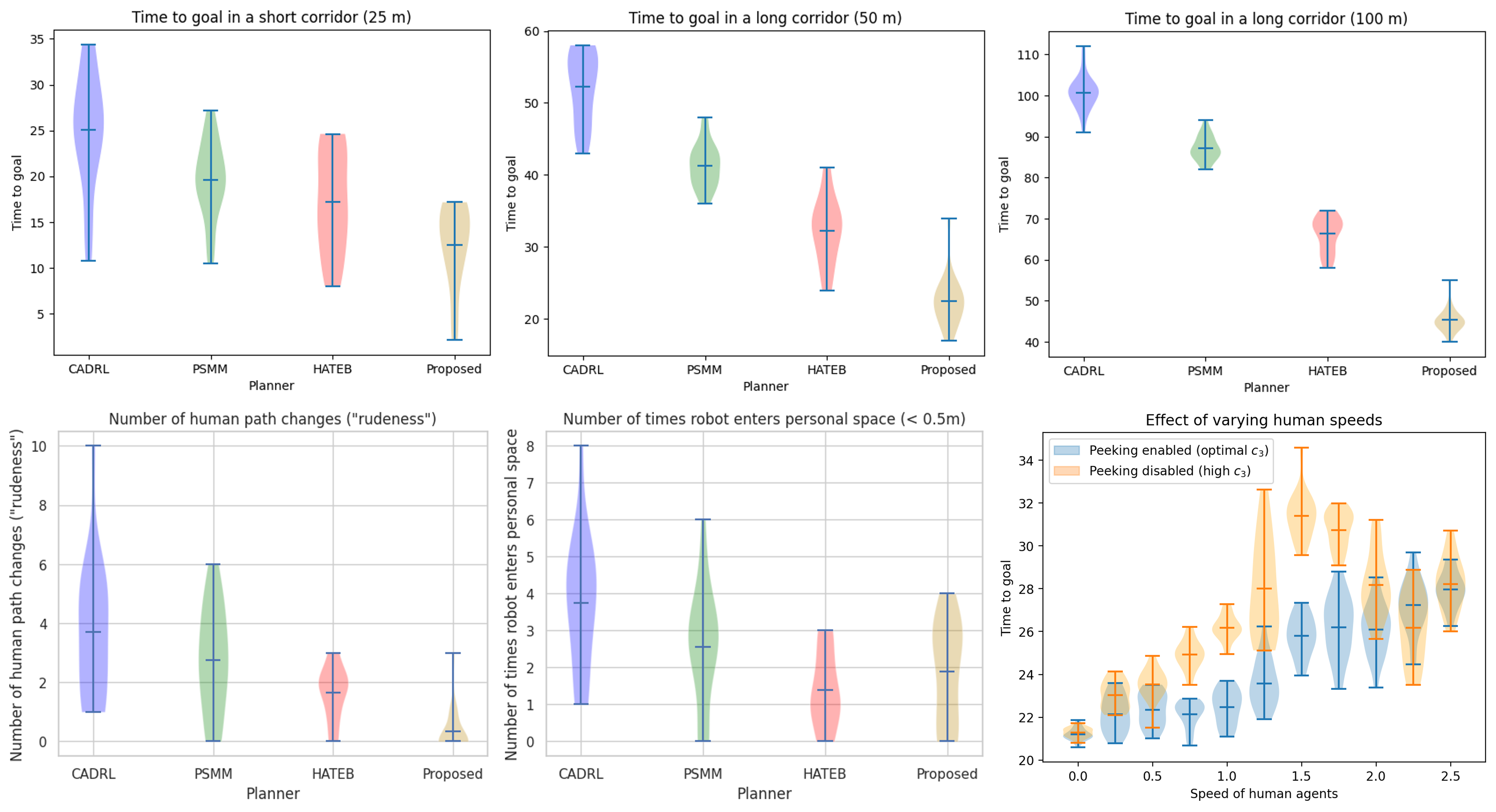}
    \caption{Trends seen when comparing the proposed planner to the selected baselines. The proposed planner outperforms commonly used state of the art planners in this uncertain social navigation scenario, with consistent results that are scalable over long distances and across a variety of human speeds.\label{fig:isaac_plots}}
\end{figure*}

The proposed pipeline is compared to three other planners: CADRL \cite{chen2017decentralized}, PSMM \cite{truong2017toward} and HATEB \cite{singamaneni2021human}. The results are shown in Fig. \ref{fig:isaac_plots}, indicate that our planner outperforms commonly used state of the art planners in this uncertain social navigation scenario, with consistent results that are scalable over long distances and across a variety of human speeds.

In particular, we make the following observations about performance (times to goal):
\begin{itemize}
    \item In short corridors of length \qty{25}{m}, our planner outperforms the baselines considered, with the average speeds showing a reasonable improvement. However, due to the short length of the corridors, the times to goal for all the planners are close and within each other's error bars.
    \item On increasing the lengths of the hallway to \qty{50}{m}, the speed advantage that our planner has is maintained, and the gap between the baselines increases.
    \item On increasing corridor length to \qty{100}{m}, the advantage that our planner possesses is clear to see, and we observe improved performance, save for a few outliers.
\end{itemize}

Next, we compare the proposed planner to the three baselines for two important metrics of social acceptability of robot motion: the number of times a human has to change their path to avoid the robot, and the number of times a robot enters a human's personal space (defined here as \qty{0.5}{m}). We demonstrate that despite the increase in performance, our planner does not appreciably sacrifice social acceptability over more conservative options. In particular, we observe that:

\begin{itemize}
    \item Our planner shows a greatly reduced number of human path changes compared to the baselines considered. In particular, HATEB is seen to be comparable to our method, with the proposed planner giving generally lower numbers of path changes than HATEB except for a few outliers. These cases are generally seen to arise due to the robot having to merge back in after a failed peek, slowing down humans behind it.
    \item The proposed pipeline is slightly worse when compared to HATEB when considering the number of times it enters a human's personal space. However, this can be attributed to our planner having to violate personal space bubbles when passing a slow human, and it is noted that such proximity usually does not make humans uncomfortable. It is therefore key to consider human path changes occurring due to the robot in tandem with personal space violations, in order to get a bigger picture.
\end{itemize}

Lastly, we analyze how robot performance varies in a \qty{50}{m} long corridor when human speeds are changed from \qty{0}{m/s} to \qty{2.5}{m/s} (a reasonable upper limit for a brisk walk by an average human). We also compare our planner's behavior to itself when peeking is turned on or off (by making $c_3$ optimal or high). We make the following observations:

\begin{itemize}
    \item With peeking enabled, when human speed increases:
    \begin{itemize}
        \item Initially, the robot travels at close to its maximum speed, with very slow or stationary humans simply being static obstacles that are easy to avoid.
        \item Beyond the static regime, the robot is still able to maintain near-optimal performance, due to the peeking and passing maneuver that our planner allows.
        \item Increasing human prediction uncertainty with increasing human speed beyond approximately \qty{1.25}{m/s} starts to force the robot to follow the human in front of it instead of attempting a peeking/passing maneuver, thereby slowing it down.
        \item Further increasing the human speed results in a "rear-end panic", where the predicted trajectories of fast humans behind the robot lead it to believe that a collision is imminent, thereby creating unnecessary "dodging" maneuvers.
    \end{itemize}
    \item With peeking disabled, when human speed increases:
    \begin{itemize}
        \item Initially, even at low human speeds, the robot is more hesitant to go around slow humans due to fear of unknown space.
        \item As human speed increases to roughly \qty{1.25}{m/s}, the predicted trajectory spread for a human in front of the robot begins to occupy the entire width of the corridor, with a high enough probability to block the risk-averse robot. This is when the robot switches from going around humans to following humans, and we see a sharp decline in performnace (increase in time to goal).
        \item Beyond \qty{1.25}{m/s}, the robot is simply following a faster human, so we observe increasing performance. (with the caveat of "rear-end panic" still present)
    \end{itemize}
\end{itemize}

These results therefore show that our proposed planner holds promise as a socially acceptable solution for high-speed robot planning in crowded, dynamic environments.



\subsection{Hardware Demonstration}

\begin{figure}[h]
    \centering
    \includegraphics[trim = 0mm 0mm 0mm 0mm, clip, width=0.45\textwidth]{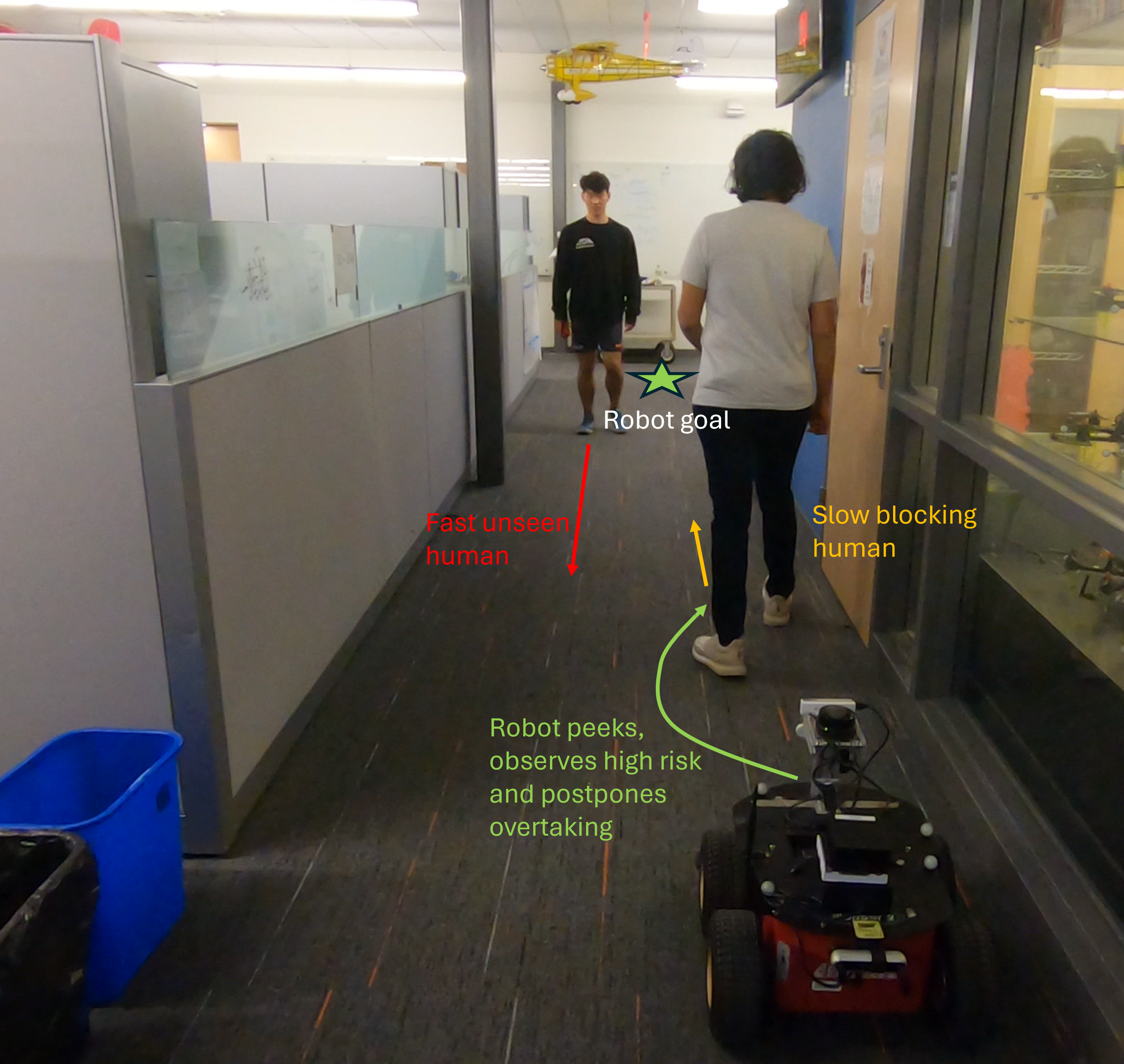}
    \caption{The proposed pipeline was shown to successfully avoid pedestrians and pass slow humans, while canceling/postponing the passing maneuver if the initial peek deems it to be too risky (such as in the case of an oncoming pedestrian).}
    \label{fig:hw_results}
\end{figure}

The pipeline described thus far was run on a differential drive ground robot platform with a max speed of \qty{1}{m/s}, in two hallway scenarios with different crowd densities. The less crowded scenario is shown in Fig. \ref{fig:hw_results}. In both scenarios, it was observed that the planner was able to follow humans in front of it at an acceptable distance without causing discomfort. Additionally, in the first scenario, the robot was seen to slow down and eventually pass a human in front of it who exhibits unexpected slowdowns. It was also observed that the robot postpones the passing maneuver if it is too risky, such as in Fig. \ref{fig:hw_results} when there is an unobserved human coming down the passing lane. However, due to the low top speed of the robot used in this early test, it was difficult to capture the robot passing humans walking at a normal pace.

\section{Conclusions and Future Work}

This paper presented an effective robot motion planner to navigate crowded and dynamic hallway scenarios at high speeds, by allowing a robot to execute "peek-and-pass" maneuvers. We demonstrated how passing a slow pedestrian can lead to significant performance gains, and offered a way to solve the "robot freezing problem". This was supported by multiple detailed simulation experiments and hardware demonstrations. 

For future work, the authors plan to do much more extensive testing with a faster robot that can reach a higher top speed of \qty{3}{m/s} or more. This is essential due to 

Additionally, the "rear-end panic" issue described at the end of Section V-B, while not likely to occur in a scenario where humans are moving at reasonable speeds, is nevertheless still an obstacle to effective and socially acceptable navigation. This is compounded by the factor that in a scenario where such decisions do matter, such as a hospital, the stakes are much higher and this property of the planner is not allowable. Addressing robot reaction to human trajectories colliding with it is a problem of two-way human-robot interactions, and we plan to draw from the rich body of work that deals with this issue.

Lastly, high speed planning entails low computation time, which has long plagued optimization-based planners such as the one presented here. We plan to look into imitation learning as an effective and proven way to accelerate our planner to handle higher speeds on smaller platforms.

\addtolength{\textheight}{-12cm}   




\section*{Acknowledgments}

This work was funded by Amazon under the Amazon Science Hub collaboration with MIT. The support and insightful comments from Zhiye Song, Nick Rober, Kota Kondo, Jeremy Cai, Andrea Tagliabue, Aneesa Sonawalla and Ne Myo Han are greatly appreciated.


\bibliographystyle{IEEEtran}
\bibliography{bibs}

\end{document}